\documentclass[11pt,a4paper]{article}
\usepackage{authblk}
\usepackage[hyperref]{acl2020}
\usepackage{times}
\usepackage{latexsym}

\usepackage{microtype}

\aclfinalcopy 

\usepackage{kotex}
\usepackage{url}
\usepackage{tabularx}
\usepackage{multirow}
\usepackage{amssymb}
\usepackage{graphicx}
\usepackage{mathtools}
\usepackage{hhline}
\usepackage{array}

\title{Integrated Eojeol Embedding for Erroneous Sentence Classification in Korean Chatbots}

\author[1,2]{ DongHyun Choi}
\author[1]{ IlNam Park}
\author[1]{ Myeong Cheol Shin}
\author[1]{ EungGyun Kim}
\author[2]{ Dong Ryeol Shin}

\affil[1]{ Kakao Enterprise, Pangyo, South Korea}
\affil[ ]{\textit {\{heuristic.c,pin.p,index.sh.jason.ng\}@kakaoenterprise.com}}
\affil[2]{Sungkyunkwan University, Suwon, South Korea}
\affil[ ]{\textit {drshin@skku.edu}}

\date{}

\begin{document}
\maketitle
\begin{abstract}
This paper attempts to analyze the Korean sentence classification system for a chatbot. Sentence classification is the task of classifying an input sentence based on predefined categories. However, spelling or space error contained in the input sentence causes problems in morphological analysis and tokenization. This paper proposes a novel approach of Integrated Eojeol (Korean syntactic word separated by space) Embedding to reduce the effect that poorly analyzed morphemes may make on sentence classification. It also proposes two noise insertion methods that further improve classification performance. Our evaluation results indicate that the proposed system classifies erroneous sentences more accurately than the baseline system by 17\%p.

\end{abstract}

\section{Introduction}
Recently, many service platforms like Amazon alexa\footnote{https://developer.amazon.com/alexa.} help third party developers create a wide range of natural language-based chatbots. Customers can access to those customized chatbots either through voice interface (employing speech recognition system), or textual interface (using chat applications or custom-built textual interface).

Natural Language Understanding for chatbot mainly consists of two components: (1) intent classification module that analyzes the user intent of an input sentence, and (2) entity extraction module that extracts entities from the input sentence. Table \ref{tab:exinputs} shows how to analyze a Korean input sentence\footnote{English translation is in the bracket.}. The first step is to classify the user intent of the input sentence, which is ``\textbf{play\_music}" in this example. Chatbot developers should predefine the set of possible intents. Once the intent is correctly classified, it is relatively easier to extract entities from the input sentence, since the classified intent helps determine the possible types of entity for the input sentence.

\begin{table}
\centering
\begin{tabular}{|l|l|} \hline
\textbf{Input}&모짜르트 클래식 틀어줘 \\&(Play Mozart's classical music) \\ \hline
\textbf{Intent}&play\_music \\ \hline
\textbf{Entities}&genre: 클래식(classic), \\&composer: 모짜르트(Mozart) \\ \hline
\end{tabular}
\caption{An example of analyzing a chatbot input sentence.}
\label{tab:exinputs}
\end{table}

When users access a chatbot through the textual interface, input sentences may occasionally contain spelling or space errors. For example, users could get confused about the spellings of the similarly pronounced words; users occasionally omit some required spaces between words; users could also make some typos.

Since Korean is an agglutinative language, its words could contain two or more morphemes to determine meanings semantically. Thus, various researchers tokenize Korean input sentences into morphemes first with a morpheme analyzer, and then the resultant morphemes are fed into a classification system \citep{Choi18,Park18_2,Hur17}. Such approaches based on morpheme embedding may work well on grammatically correct sentences but show poor performances on erroneous inputs, as shown in section \ref{sec:er}. Errors in input sentences are likely to cause problems in the process of morphological analysis. Occasionally, morphemes with significant meaning are miscategorized as meaningless morphemes due to spelling errors. Table \ref{tab:wrongmorp} briefly illustrates the case mentioned above. An important clue ``파케스트", the typo of a Korean morpheme ``팟케스트(podcast)", is separated into two meaningless morphemes due to spelling errors.

In this paper, a novel approach of Integrated Eojeol Embedding (\textbf{IEE}) is proposed to handle the classification problem. \textit{Eojeol} is a Korean term meaning a word. In this paper, \textit{Eojeol} is operationally defined as a sequence of Korean characters, separated by spaces. Detailed examples are given in section \ref{sec:korexp}.

The main idea of \textbf{IEE} is to feed the Eojeol embedding vectors into the sentence classification network, instead of morphemes or other subword units embedding vectors. In the case of an Eojeol $\textbf{w}$, subword unit-based Eojeol embedding vectors are calculated first based on different subword units of \textbf{w}, and the resultant vectors are integrated to form a single \textbf{IEE} vector. By doing so, the algorithm could significantly reduce the effect of incorrect subword unit tokenization results caused by spelling or other errors, while maintaining the benefits from the pre-trained subword unit embedding vectors such as GloVe \citep{pennington14} or BPEmb \citep{heinzerling2018bpemb}. 

\begin{table}
\centering
\begin{tabularx}{\linewidth}{|l|X|} \hline
\textbf{With Typo}& 딴 파케스트로 플레이\\ \hline
\textbf{Morphemes}&딴(another) / 파케(podca) / 스트로(to st) / 플레이(play)\\ \hline
\textbf{Corrected} & 딴 팟케스트로 플레이 \\ \hline
\textbf{Morphemes}&딴(another) / \textbf{팟케스트(podcast)} /로(to) / 플레이(play) \\ \hline
\textbf{Meaning}&Play another podcast \\ \hline
\end{tabularx}
\caption{An example of miscategorized morphemes due to a typo in the sentence.}
\label{tab:wrongmorp}
\end{table}

Also, two noise insertion methods called \textit{Jamo dropout} and \textit{space-missing sentence generation} are proposed to automatically insert noisy data into the training corpus and enhance the performance of the proposed \textbf{IEE} approach. The proposed system outperforms the baseline system by over 18\%p on the erroneous sentence classification task, in terms of sentence accuracy.

In Section \ref{sec:related}, related works are briefly reviewed. Section \ref{sec:alg} describes the \textbf{IEE} approach and the two proposed noise insertion methods in more detail. Section \ref{sec:exp} shows evaluation results of the proposed system. Finally, conclusions and future works are given in Section \ref{sec:con}.

\section{Related Works}
\label{sec:related}
There exists a wide range of previous works for English sentence classification. \citet{Kim14} employed CNN with max-pooling for sentence classification, \citet{Bowman15} used BiLSTMs to get the sentence embeddings for natural language inference tasks, and \citet{Zhou15} tried to combine the LSTM with CNN. Recent works such as \citet{Im17}  or \citet{Yoon18} tried to explore the self-attention mechanism for sentence encoding. However, to the best of our knowledge, the classification task of erroneous sentences received much less attention.

This paper mainly focuses on integrating multiple embeddings. \citet{Yin16} also considered the idea of integrating multiple embeddings; they considered many different embeddings as multiple channel inputs and extended the algorithm of \citet{Kim14} to handle the multi-channel inputs. This paper mainly differs from their work in that we attempt to generalize their approach. Unlike them, \textbf{IEE} does not require the input embeddings to have the same subword tokenizations while integrating various embeddings.

Besides, \citet{Choi18} proposed morpheme-based Korean GloVe word embedding vectors along with Korean word analogy corpus to evaluate them, while GloVe \citep{pennington14} is a pre-trained embedding vector set using unstructured text. \citet{Choi18} also used the trained Korean GloVe embedding vectors with the algorithm of \citet{Kim14} to train the Korean sentence classifier for Korean chatbots. \citet{Park18} focused on sentence classification problem for sentences with spacing errors, but other types of errors such as typo are ignored. 

In this paper, the proposed \textbf{IEE} vectors are fed into the network proposed in \citet{Choi18}. The overall system performance is compared against the original sentence classification system of \citet{Choi18} and \citet{Kim14}, to clarify the effect of \textbf{IEE} vectors.

\section{Classifying Erroneous Korean Sentences}
\label{sec:alg}
In this section, an algorithm to correctly classify erroneous Korean sentences is proposed.

\subsection{Brief Introduction to Korean Word and Syntactic Structures}
\label{sec:korexp}
In this subsection, the structures of Korean words and sentences are briefly described to help to understand this paper. \textit{Eojeol} is defined as a sequence of Korean characters, distinguished by spaces. A given input sentence \textbf{s} is tokenized using spaces to get its constituent Eojeol list $\{w_1, w_2, ..., w_s\}$. An Eojeol contains one or more morphemes, which can be extracted using a Korean morphological analyzer. Table \ref{tab:eojeol} shows the Eojeols and morphemes of the example sentence from Table \ref{tab:wrongmorp}.

\begin{table}
\centering
\begin{tabular}{|l|l|} \hline
\textbf{Sentence}& 딴 팟케스트로 플레이 \\ \hline
\textbf{Eojeols}&딴 / 팟케스트로 / 플레이 \\ \hline
\textbf{Morphemes}& 딴 / 팟케스트 / 로 / 플레이 \\ \hline
\end{tabular}
\caption{An example of Eojeols and morphemes of the sentence from Table \ref{tab:wrongmorp}.}

\label{tab:eojeol}
\end{table}

\begin{table}
\begin{tabularx}{\linewidth}{|c|X|} \hline
\textbf{Placement}&\multicolumn{1}{|c|}{\textbf{Possible Candidates}} \\ \hline
\textit{initial}&ㄱ, ㄴ, ㄷ, ㄹ, ㅁ, ㅂ, ㅅ, ㅇ, ㅈ, ㅊ, ㅋ, ㅌ, ㅍ, ㅎ, ㄲ, ㄸ, ㅃ, ㅆ, ㅉ\\ \hline
\textit{medial}&ㅏ, ㅐ, ㅑ, ㅒ, ㅓ, ㅔ, ㅕ, ㅖ, ㅗ, ㅘ, ㅙ, ㅚ,  ㅛ, ㅜ, ㅝ, ㅞ, ㅟ, ㅠ, ㅡ,  ㅢ, ㅣ\\ \hline
\textit{final}&ㄱ, ㄴ, ㄷ, ㄹ, ㅁ, ㅂ, ㅅ, ㅇ, ㅈ, ㅊ, ㅋ, ㅌ, ㅍ, ㅎ, ㄲ, ㅆ, ㄳ, ㄵ, ㄶ, ㄺ, ㄻ, ㄼ, ㄽ, ㄾ, ㄿ, ㅀ, ㅄ\\ \hline
\end{tabularx}
\caption{Possible candidate Jamos for each Korean character component.}
\label{tab:KorChars}
\end{table}

\begin{table}
\begin{tabular}{|c|c|c|c|} \hline
\textbf{Korean Character}&\textit{\textbf{Initial}} &\textit{\textbf{Medial}}&\textit{\textbf{Final}}\\ \hline
케&ㅋ&ㅔ&\\ \hline
팟&ㅍ&ㅏ&ㅅ\\ \hline
\end{tabular}
\caption{Examples of Korean characters and their constituent Jamos.}
\label{tab:KorCharComp}
\end{table}

A Korean character consists of two or three Korean alphabets or \textit{Jamo}s; a consonant called the \textit{initial} (or \textit{choseong} in Korean), a vowel called the \textit{medial} (or \textit{jungseong}), and optionally another consonant called the \textit{final} (or \textit{jongseong}). 19 consonants are used as the \textit{initial}, 21 vowels as the \textit{medial}, and 27 consonants as the \textit{final}. Table \ref{tab:KorChars} shows the list of possible Jamo candidates for each placement. Theoretically, there can be 11,172($=19\times21\times(27+1)$, considering the cases without the \textit{final}s) different kinds of Korean characters in total, but only 2,000 to 3,000 characters are being used in the real world.

Table \ref{tab:KorCharComp} shows a few examples of Korean characters having constituent Jamos. The first example has no \textit{final} but still makes valid Korean characters. There exist 51 Korean Jamos, 30 consonants, and 21 vowels in total, except duplications.  

\subsection{Integrated Eojeol Embedding}
\label{sec:uee}

\begin{figure}
\centering
\includegraphics[width=\linewidth]{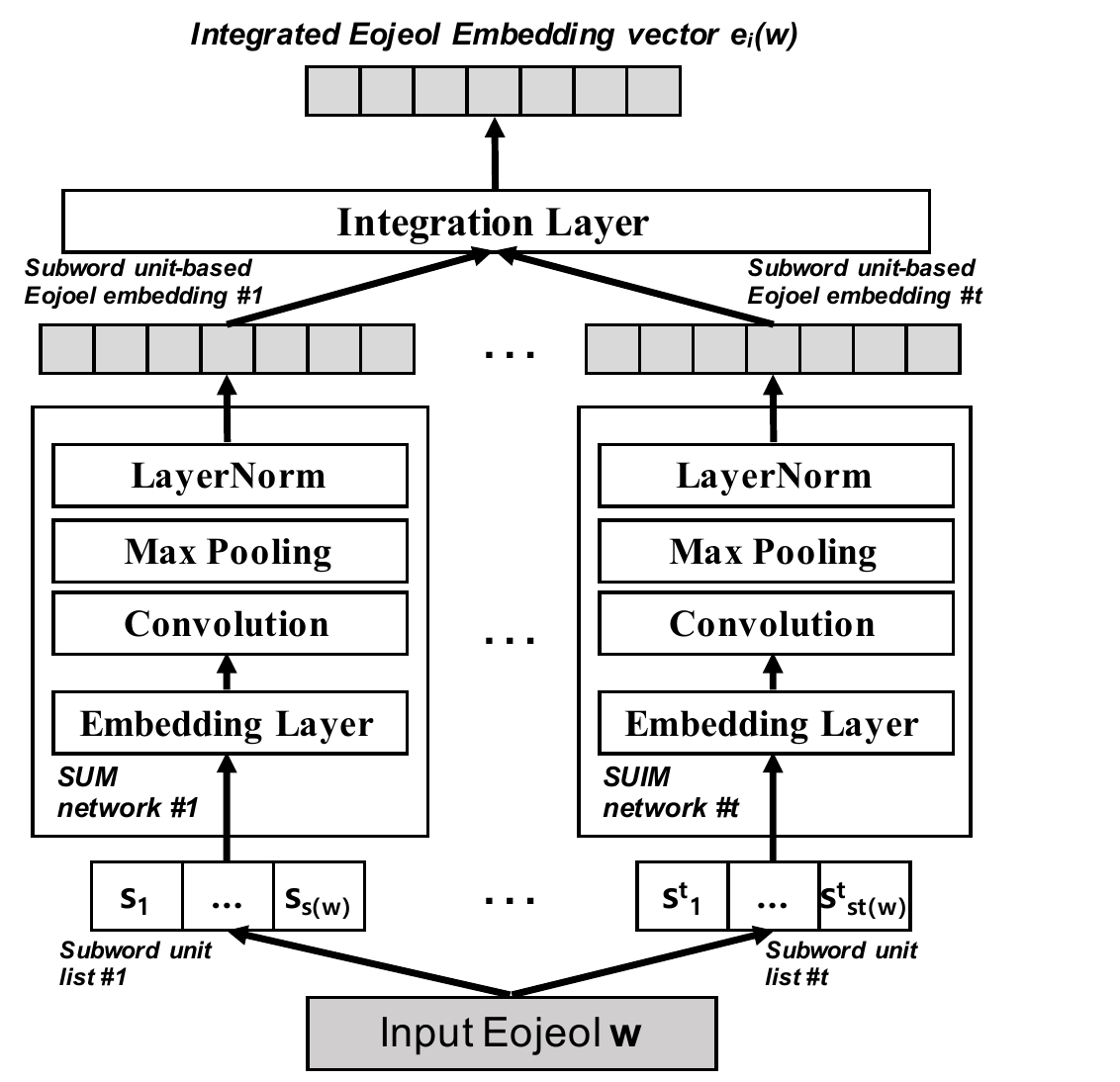}
\caption{Network architecture to get the proposed Integrated Eojeol Embedding.}
\label{fig:uee}
\end{figure}

Figure \ref{fig:uee} illustrates the network architecture to calculate an \textbf{IEE} vector for a given Eojeol \textbf{w}. For an Eojeol \textbf{w}, \textbf{t} types of its subword unit lists are generated first. In this paper, four types of subword unit lists are considered: Jamo list $L_J(\textbf{w})=\{j_1, ...,j_{l_j(\textbf{w})}\}$, character list $L_C(\textbf{w})=\{c_1, ...,c_{l_c(\textbf{w})}\}$,  byte-pair encoding (BPE) subunit list $L_B(\textbf{w})=\{b_1, ...,b_{l_b(\textbf{w})}\}$, and morpheme list $L_M(\textbf{w})=\{m_1, ...,m_{l_m(\textbf{w})}\}$, where $l_j(\textbf{w})$, $l_c(\textbf{w})$, $l_b(\textbf{w})$ and $l_m(\textbf{w})$ are the lengths of the lists. Table \ref{tab:subunitexample} shows the subword unit lists of an Eojeol ``팟케스트로".

\begin{table}[hbt!]
\begin{tabular}{|l|l|} \hline
\textbf{Unit} & \textbf{Subword Unit List}\\ \hline
\textbf{Jamo}&ㅍ/ㅏ/ㅅ/ㅋ/ㅔ/ㅅ/ㅡ/ㅌ/ㅡ/ㄹ/ㅗ \\ \hline
\textbf{Character}&팟 / 케 / 스 / 트 / 로 \\ \hline
\textbf{BPE unit}& \_ / 팟 / 케 / 스트로 \\ \hline
\textbf{Morpheme}&팟케스트 / 로 \\ \hline
\end{tabular}
\caption{Subword unit lists of an Eojeol "팟캐스트로", which is retrieved from Table \ref{tab:eojeol}.}
\label{tab:subunitexample}
\end{table}

Each subword unit list is then fed into the subword unit merge (SUM) network. For a subword unit list $L_S(\textbf{w})$, the SUM network first converts each list item into its corresponding embedding vector to get an embedding matrix $E_S(\textbf{w})$. Afterward, one-dimensional depthwise separable convolutions \citep{Chollet17} with kernel sizes $k = 2,3,4,5$, and filter size $F$ are applied on $E_S(\textbf{w})$; the results are followed by max-pooling and layernorm  \citep{Ba16}. The result of the SUM network is the subword unit-based Eojeol embedding vector $e_S(\textbf{w}) \in \mathbb{R}^{4F}$. The Integrated Eojeol Embedding vector $e_i(\textbf{w})$ is then calculated by integrating all the subword unit-based Eojeol embedding vectors. For subword unit types $T=\{s_1, ..., s_t\}$, three different algorithms to construct the \textbf{IEE} are proposed.

\begin{itemize}
\item \textbf{IEE by Concatenation}. All the subword unit-based Eojeol embedding vectors are concatenated to form one \textbf{IEE} vector: $e_i(\textbf{w}) = [ e_{s_1}(\textbf{w}); ...; e_{s_t}(\textbf{w}) ] \in \mathbb{R}^{4tF}$.
\item \textbf{IEE by Weighted Sum}. Weights of the subword unit-based Eojeol embedding vectors are trained, and \textbf{IEE} vector is defined as the weighted sum of the subword unit-based Eojeol embedding vectors. More precisely, $e_i(\textbf{w}) = W \cdot [ e_{s_1}(\textbf{w}), ..., e_{s_t}(\textbf{w}) ] \in \mathbb{R}^{4F}$, where $ W = M \cdot \tanh( \sum_{s \in T} M_{s}e_{s}(\textbf{w})) \in \mathbb{R}^{t}$, $M \in \mathbb{R}^{t \times 4F}$ and $M_s \in \mathbb{R}^{4F \times 4F}$.

\item \textbf{IEE by Max Pooling}. Set the $j$-th element of \textbf{IEE} as the maximal value of $j$-th elements of the subword unit-based Eojeol embedding vectors: $e_i(\textbf{w}) = \max[ e_{s_1}(\textbf{w}), ..., e_{s_t}(\textbf{w}) ] \in \mathbb{R}^{4F}$.
\end{itemize}

\subsection{IEE-based Sentence Classification}
\label{sec:overall}
\begin{figure}[h]
\centering
\includegraphics[width=0.8\linewidth]{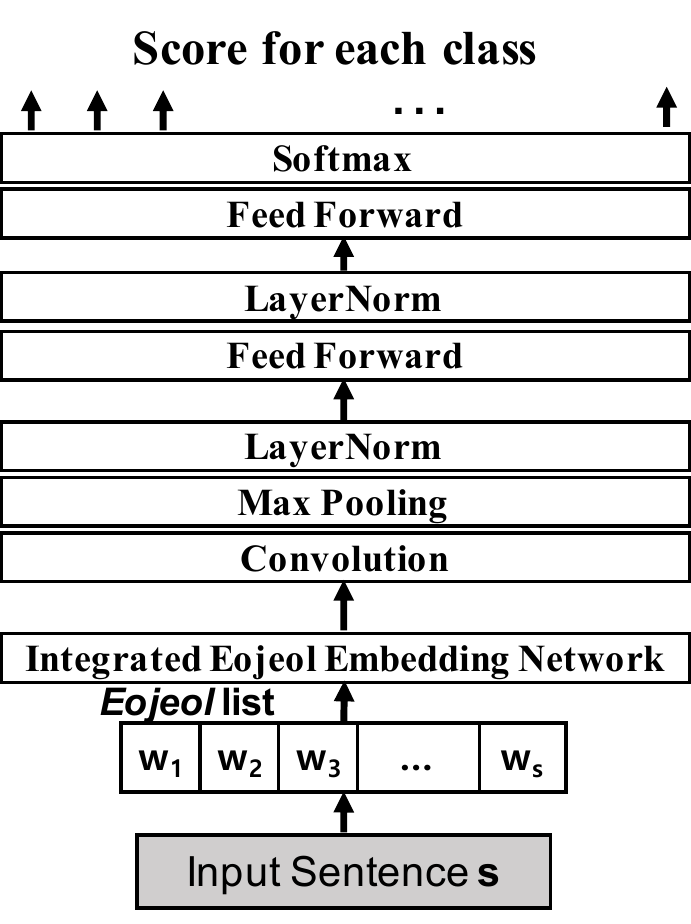}
\caption{Network architecture for sentence classification}
\label{fig:overall}
\end{figure}

Figure \ref{fig:overall} illustrates the network architecture for sentence classification that uses the proposed \textbf{IEE} approach. A given Korean sentence \textbf{s} is considered as the list of Eojeols $\{w_1, w_2, ..., w_s\}$. For each Eojeol $w_i$, $e_i(w_i)$ is firstly calculated with the \textbf{IEE} network proposed in \ref{sec:uee} to get an \textbf{IEE} matrix $E(\textbf{s}) = [e_i(w_1),e_i(w_2), ..., e_i(w_s)] \in \mathbb{R}^{s \times d(I)}$. In the formula, $d(I)$ is the dimension of the \textbf{IEE} vector; $\mathbb{R}^{4tF}$ for \textbf{IEE by Concatenation} and $\mathbb{R}^{4F}$ otherwise.

Once $E(\textbf{s})$ is calculated, depthwise separable convolutions with kernel sizes $k^\prime = 1, 2, 3, 4, 5$ are applied on the matrix, and the results are followed by max-pooling and layernorm. Finally, the two fully connected feed-forward layers are applied to get the final output score vector $O \in \mathbb{R}^c$, while $c$ is the number of possible classes. The overall network architecture is a variation of the network architecture proposed in \citet{Choi18}, with its input replaced by the newly proposed \textbf{IEE} vectors.

\subsection{Noise Insertion Methods to Improve the Integrated Eojeol Embedding}
\label{sec:supmethod}
In this subsection, two noise insertion methods that further improve the performance of the \textbf{IEE} vectors are proposed. The first method, called \textit{Jamo dropout}, masks Jamos during the training phase with Jamo dropout probability or \textit{jdp}. Instead of masking some elements of the input embedding vector as regular dropout \citep{Srivastava14} does, \textit{Jamo dropout} masks the whole Jamo embedding vector. Also, if any subword unit of different types contains masked Jamos, the embedding vector of that subword unit is masked. The method has two expected roles. First, it introduces additional noise during the training phase to make the trained classifier work better on noisy inputs. Secondly, by masking other subword units that contain masked Jamos allow the system to learn how to focus on the Jamo embeddings when the input subword unit is unknown in trained subword unit vocabulary. 

Compared to the morpheme-embedding-based approach, the Eojeol-embedding-based approach is expected to show poor performance on sentences without spaces, since Eojeols are gathered by tokenizing the input sentence with spaces. Therefore, the second noise insertion method called \textit{space-missing sentence generation} is introduced to resolve such impediments. This method aims to process input sentences without necessary spaces by automatically adding the inputs into the training corpus. More precisely, sentences are randomly chosen from the training corpus with the probability \textit{msp} (missing space probability), and all the spaces in the chosen sentences are removed. The space-removed sentences are then inserted into the training corpus. The \textit{space-missing sentence generation} method is applied only once before the training phase. 

\section{Experiments}
\label{sec:exp}
In this section, experimental settings and evaluation results are described.

\subsection{Corpus} 

The intent classification corpus proposed in \citet{Choi18} contains 127,322 manually-annotated, grammatically correct Korean sentences with 48 intents. Examples are \textit{weather} (Ex: 오늘 날씨 어때 How is the weather today), \textit{fortune} (Ex: 내일 운세 알려줘 Tell me the tomorrow’s fortune), \textit{music} (Ex: 음악 틀어줘 Play music) and \textit{podcast} (Ex: 팟캐스트 틀어줘 Play podcast). Sentences representing each intent are randomly divided into 8:1:1 ratio for train, validation, and test dataset. The test dataset here is called as the \textbf{WF} (Well-Formed) corpus throughout the paper and consists of 12,711 sentences.

The \textbf{KM} corpus standing for Korean Mis-spelling is manually annotated to measure the system performance on erroneous input sentences. The \textbf{KM} corpus is annotated with 46 intents, with two intents fewer than the \textbf{WF} corpus. The removed two intents are \textit{OOD} and \textit{Common}; \textit{OOD} is the intent for meaningless sentences, and \textit{Common} is the intent for short answers, such as ``응 yes.” The sentences for the \textit{Common} intent are too short of recovering their meanings from errors.

For 46 intents of the \textbf{KM} corpus, two annotators are asked to create sentences for 23 intents each. For each intent, an annotator is firstly asked to create 45 sentences without errors. After then, the annotator is asked to insert errors on the created sentences. The error insertion guideline is given in Table \ref{tbl:guideline}. As a result, the \textbf{KM} corpus contains 2,070 erroneous sentences in total, and it is used only for testing.
\begin{table}
 \centering
\begin{tabularx}{\linewidth}{|X|} \hline 
\textbf{RULE 1.} For each sentence, insert one or more errors. Some recommendations are: \\
- Remove / Duplicate a Jamo. \\
- Swap the order of two or more Jamos. \\
- Replace a Jamo with similarly pronounced. \\
- Replace a Jamo with nearly located on keyboard. \\ \hline
\textbf{RULE 2.} The erroneous sentences should be understandable. \\
- Each annotator tries to classify other's works. Misclassified sentences are reworked. \\ \hline
\textbf{RULE 3.} For 45 sentences of each intent, 25 sentences should contain only valid characters, and 20 sentences should contain one or more invalid characters. \\ 
- ``Invalid" Korean characters are defined as those without \textit{initial} or \textit{medial}. \\ 
\hline
\end{tabularx}
 \caption{\textbf{KM} corpus annotation guidelines}

\label{tbl:guideline}
\end{table}

Also, a space-missing (\textbf{SM}) test corpus is generated based on the \textbf{WF} and \textbf{KM} corpus to evaluate the system performance on sentences without necessary spaces. The corpus consists of sentences from the other test corpus, but the spaces between words are randomly removed with the probability of 0.5, and at least one space is removed from each original sentence. The \textbf{SM} corpus contains 14,781 sentences and also is used only for testing.

\subsection{Experimental Setup}
Sentence accuracy is used as a criterion to measure the performances of sentence classification systems on each corpus. The sentence accuracy or \textit{SA} is defined as follows:

\begin{align}
SA=\frac{\#\ correct\ sentences}{\#\ total\ sentences}
\end{align}

Throughout the experiments, the value of \textit{jdp} is set to 0.05, and \textit{msp} is set to 0.4. Those values are chosen through grid search with ranges $jdp=\{0.00, 0.05, 0.10, 0.15, 0.20\}$ and $msp=\{0.0, 0.1, 0.2, 0.3, 0.4\}$. For each experiment configuration, three training and test procedures are carried out, and the average of three test results on each test corpus is presented as the final system performance. 

ADAM optimizer \citep{Kingma14} with learning rate warm-up scheme is applied. The learning rate increases from 0.0 to 0.1 in the first three epochs, and exponential learning rate decay with a decay rate of 0.75 is applied after five epochs of training. On each epoch, the trained classifier is evaluated against the validation dataset, and the training stops when the validation accuracy is not improved for four consequent training epochs. Minibatch size is set to 128. Dropout \citep{Srivastava14} is applied between layers, with a rate of 0.1. 

Korean GloVe embedding vectors proposed by \citet{Choi18} are used for the morpheme embedding vector. The dimension of GloVe embedding is 300. Also, BPEmb proposed in \citet{heinzerling2018bpemb} is used for BPE unit embedding. For BPEmb, vector dimension and vocabulary size are experimentally set to 300 and 25,000, respectively. The values of morpheme embedding vectors and BPE unit embedding vectors are fixed during training. Meanwhile, Jamo embedding vectors and character embedding vectors are trained altogether with network parameters. The dimensions of Jamo and character embedding vectors are both set to 300, and the convolutional filter size $F$ is set to 128.

KHAIII\footnote{https://github.com/kakao/khaii} Korean morphological analyzer is used to analyze the morphemes. The network is implemented with tensorflow\footnote{https://www.tensorflow.org}. 

\subsection{Evaluation Results}
\label{sec:er}
In this subsection, the system configurations are notated as follows: $\textbf{IEE}_\textbf{C}$ represents \textbf{IEE by Concatenation}, $\textbf{IEE}_\textbf{W}$ represents \textbf{IEE by Weighted Sum}, and $\textbf{IEE}_\textbf{M}$ represents \textbf{IEE by Max-pooling}, as described in subsection \ref{sec:uee}. Also, the applied subword units are represented in a square bracket;  $m$ for morpheme, $b$ for BPE unit, $j$ for Jamo, and $c$ for character. The use of noise insertion methods are represented after \textbf{+} sign;  \textbf{SG} represents the application of \textit{space-missing corpus generation}, and \textbf{JD} represents the application of \textit{Jamo dropout}, and \textbf{ALL} means both.

\begin{table} 
 \centering
\begin{tabular}{|l|c|c|c|} \hline 
\textbf{System}&\textbf{WF}&\textbf{KM}&\textbf{SM} \\ \hline
\textbf{Kim-1 \shortcite{Kim14}}&87.92&44.53&33.10 \\
\textbf{Kim-2 \shortcite{Kim14}}&95.66&46.53&85.19 \\
\textbf{Choi et al. \shortcite{Choi18}}&96.91&50.29&87.46 \\ 
\textbf{M-BERT \shortcite{bert}}  & 97.00 &51.09&77.40 \\ \hhline{|====|}
$\textbf{IEE}_\textbf{C}[mj]\textbf{+ALL}$ &97.32&\textbf{68.33}&91.18\\ 
$\textbf{IEE}_\textbf{C}[mjc]\textbf{+ALL}$ &97.27&68.10&\textbf{91.36}\\ 
$\textbf{IEE}_\textbf{C}[mbj]\textbf{+ALL}$ &\textbf{97.47}&65.23&91.11\\ \hline
\end{tabular}
 \caption{Comparison of the baseline systems and the proposed \textbf{IEE} approach}

\label{tbl:exp_compare}
\end{table}

Table \ref{tbl:exp_compare} compares the performances of the proposed system and the baseline systems. The first and second baseline systems are based on the algorithm proposed by \citet{Kim14}. Source code is retrieved from the author's repository and modified to get the Korean sentences as its input. Baseline \textbf{Kim-1} gets the Eojeol list of an input sentence as its input, and baseline \textbf{Kim-2} instead gets the morpheme list of the given sentence as its input. The third baseline system is a system proposed in \citet{Choi18}. It also receives the list of Korean morphemes as its input. The fourth baseline system \textbf{M-BERT} is the multilingual version of BERT \citep{bert}. We downloaded the pretrained model from the author's repository, and fine-tuned it for Korean sentence classification. For the proposed approach, the performances of three different system configurations are selected and presented.

As can be observed from the table, the performance on the \textbf{KM} corpus improved over 17\%p compared to the baseline systems. The proposed system outperforms the baseline systems on the \textbf{WF} and the \textbf{SM} corpus by integrating information from different types of subword units.

Another set of experiments is carried out to configure the effect of each noise insertion method and to compare the three proposed \textbf{IEE} approaches. All four subword unit types are used throughout the experiments, and evaluation results are presented in Table \ref{tbl:exp_iee}.  As can be observed from the table, using the \textbf{IEE} approaches with the two proposed noise insertion methods dramatically improves the system performance on the erroneous sentence classification task. The performance on the \textbf{KM} corpus increased about 14 to 15\%p on every proposed \textbf{IEE} approach. 

As expected in section \ref{sec:supmethod}, the \textbf{IEE} approaches show relatively low performance on the \textbf{SM} corpus without the application of \textbf{SG}, compared to the baseline systems presented in Table \ref{tbl:exp_compare}. However, the evaluation result suggests that a decrease in performance could be efficiently handled by applying the \textbf{SG} method. The performance of $\textbf{IEE}_\textbf{C}[mbjc]\textbf{+SG}$  system on \textbf{SM} corpus reaches up to 89.40\%, which is about 2\%p higher than the \textbf{SM} corpus performances in the baseline systems. Among the three proposed \textbf{IEE} approaches, $\textbf{IEE}_\textbf{C}$ performed better than the other two approaches in most cases. $\textbf{IEE}_\textbf{M}$ performed slightly better than the $\textbf{IEE}_\textbf{C}$ on the \textbf{KM} corpus but showed low performance on the \textbf{WF} corpus. The performance of $\textbf{IEE}_\textbf{W}$ on the \textbf{KM} corpus was much lower than the other two approaches.
    
\begin{table}
\centering
\begin{tabular}{|l|c|c|c|} \hline 
\textbf{System}&\textbf{WF}&\textbf{KM}&\textbf{SM} \\ \hline
$\textbf{IEE}_\textbf{C}[mbjc]$ &\textbf{97.23}&50.00&73.49 \\
$\textbf{IEE}_\textbf{M}[mbjc]$ &96.99&\textbf{51.39}&\textbf{75.21} \\
$\textbf{IEE}_\textbf{W}[mbjc]$ &97.11&49.18&73.70 \\ \hline

$\textbf{IEE}_\textbf{C}[mbjc]\textbf{+JD}$ &\textbf{97.41}&\textbf{64.83}&75.92 \\
$\textbf{IEE}_\textbf{M}[mbjc]\textbf{+JD}$ &96.98&63.69&\textbf{77.50} \\
$\textbf{IEE}_\textbf{W}[mbjc]\textbf{+JD}$ &97.36&63.06&75.68 \\ \hline

$\textbf{IEE}_\textbf{C}[mbjc]\textbf{+SG}$ &\textbf{97.26}&54.07&\textbf{89.40} \\
$\textbf{IEE}_\textbf{M}[mbjc]\textbf{+SG}$ &96.95&\textbf{55.80}&89.05 \\
$\textbf{IEE}_\textbf{W}[mbjc]\textbf{+SG}$ &97.13&54.38&89.32 \\ \hline

$\textbf{IEE}_\textbf{C}[mbjc]\textbf{+ALL}$ &\textbf{97.34}&65.03&91.01 \\ 
$\textbf{IEE}_\textbf{M}[mbjc]\textbf{+ALL}$ &97.04&\textbf{65.30}&90.95 \\ 
$\textbf{IEE}_\textbf{W}[mbjc]\textbf{+ALL}$ &\textbf{97.34}&62.93&\textbf{91.14} \\ \hline

\end{tabular}
\caption{Comparison results between the three proposed \textbf{IEE} approaches with the two noise insertion methods. }

\label{tbl:exp_iee}
\end{table}
  
 To distinguish the contribution of the Eojeol-based approach from the contributions of the two noise insertion methods, two subwords unit-based integrated embedding approaches are newly defined. Integrated morpheme embedding (\textbf{IME}) approach creates an integrated morpheme embedding vector for each morpheme by integrating the Jamo-based morpheme embedding vector and the character-based morpheme embedding vector with the pre-trained morpheme embedding vector. The integrated BPE unit embedding (\textbf{IBE}) approach creates an integrated BPE embedding vector for each BPE unit in the same manner. The two approaches are the same as the \textbf{IEE} approach, except that \textbf{IME} and \textbf{IBE} feed the morpheme embedding vector and BPE unit embedding vector respectively while the \textbf{IEE} calculates and feeds the Eojeol embedding vector into the network in figure \ref{fig:overall}. $[mjc]$ is used to compare \textbf{IME} and \textbf{IEE} subword unit settings, since the BPE embeddings cannot be integrated into \textbf{IME}. To compare \textbf{IBE} and \textbf{IEE} subword unit settings, $[bjc]$ is used for the same reason. For the integration method, the concatenation is only considered in the experiments.

Table \ref{tbl:exp_eeme} shows the comparison result. The two noise insertion methods worked well on $\textbf{IME}_\textbf{C}$ and $\textbf{IBE}_\textbf{C}$; however, for the \textbf{KM} corpus, the performance of the Eojeol-based embedding approach is 3\%p higher than that of morpheme-based or BPE-based approaches. The result shows that the Eojeol-based embedding approach handles with the subword unit analysis errors efficiently, compared to the subword unit-based integrated embedding approaches such as \textbf{IME} or \textbf{IBE}.

\begin{table}
 \centering
\begin{tabular}{|l|c|c|c|} \hline 
\textbf{System}&\textbf{WF}&\textbf{KM}&\textbf{SM} \\ \hline
$\textbf{IME}_\textbf{C}[mjc]$ &96.93&51.18&87.76 \\
$\textbf{IME}_\textbf{C}[mjc]$\textbf{+JD} &97.23&62.35&90.00 \\
$\textbf{IME}_\textbf{C}[mjc]$\textbf{+SG} &96.98&56.30&90.33 \\
$\textbf{IME}_\textbf{C}[mjc]$\textbf{+ALL}  &97.28&65.09&\textbf{91.66} \\ \hline
$\textbf{IEE}_\textbf{C}[mjc]$\textbf{+ALL} &\textbf{97.32}&\textbf{68.10}&91.36 \\ \hhline{|====|}
$\textbf{IBE}_\textbf{C}[bjc]$ &96.45&54.83&79.76 \\
$\textbf{IBE}_\textbf{C}[bjc]$\textbf{+JD} &96.69&57.79&83.02 \\
$\textbf{IBE}_\textbf{C}[bjc]$\textbf{+SG}  &96.26&55.96&89.70 \\
$\textbf{IBE}_\textbf{C}[bjc]$\textbf{+ALL} &96.54&61.43&\textbf{90.89} \\ \hline
$\textbf{IEE}_\textbf{C}[bjc]$\textbf{+ALL} &\textbf{96.74}&\textbf{64.97}&90.37 \\ \hline
\end{tabular}
 \caption{The comparison results of Eojeol based approach and other subword unit-based integrated embedding approaches.}

\label{tbl:exp_eeme}
\end{table}

\begin{table}[hbt!]
\centering
\begin{tabular}{|l|c|c|c|} \hline 
\textbf{System}&\textbf{WF}&\textbf{KM}&\textbf{SM} \\ \hline
$\textbf{IEE}_\textbf{C}[m]\textbf{+SG}$ &97.04&49.58&86.42 \\
$\textbf{IEE}_\textbf{C}[b]\textbf{+SG}$ &96.27&55.29&87.73 \\
$\textbf{IEE}_\textbf{C}[c]\textbf{+SG}$ &95.31&\textbf{60.37}&88.58 \\ 
$\textbf{IEE}_\textbf{C}[mb]\textbf{+SG}$&\textbf{97.38}&51.34&88.70 \\
$\textbf{IEE}_\textbf{C}[mc]\textbf{+SG}$ &97.17&54.17&89.13 \\
$\textbf{IEE}_\textbf{C}[bc]\textbf{+SG}$ &96.40&59.68&89.20 \\ 
$\textbf{IEE}_\textbf{C}[mbc]\textbf{+SG}$ &97.20&55.88&\textbf{89.26} \\ \hline
$\textbf{IEE}_\textbf{C}[j]\textbf{+ALL}$ &95.19&\textbf{69.16}&89.09 \\
$\textbf{IEE}_\textbf{C}[mj]\textbf{+ALL}$ &97.32&68.33&91.18 \\
$\textbf{IEE}_\textbf{C}[bj]\textbf{+ALL}$ &96.60&64.04&90.13 \\
$\textbf{IEE}_\textbf{C}[jc]\textbf{+ALL}$ &95.89&68.15&90.22 \\ 
$\textbf{IEE}_\textbf{C}[mbj]\textbf{+ALL}$&\textbf{97.47}&65.23&91.11 \\ 
$\textbf{IEE}_\textbf{C}[mjc]\textbf{+ALL}$ &97.27&68.10&\textbf{91.36}\\
$\textbf{IEE}_\textbf{C}[bjc]\textbf{+ALL}$ &96.74&64.97&90.37 \\ 
$\textbf{IEE}_\textbf{C}[mbjc]\textbf{+ALL}$ &97.34&65.03&91.01 \\ \hline
\end{tabular}
\caption{Experiment results on the effect of each subword unit embedding}

\label{tbl:subunit}
\end{table}

Finally, experiments are conducted on $\textbf{IEE}_\textbf{C}$ system with various subword unit configurations to figure out the effect of each subword unit on system performance. Table \ref{tbl:subunit} shows the comparison result. \textbf{SG} noise insertion method is applied only to the systems that have no Jamo subword units, while the two noise insertion methods are applied to other cases. The evaluation results are compared between the two different groups; one with the Jamo subword unit and one without it.

Several interesting facts can be observed from the table. First, the application of the Jamo subword unit dramatically improves the system performance on the \textbf{KM} corpus. The performance of the $\textbf{IEE}_\textbf{C}[j]\textbf{+ALL}$ system on the \textbf{KM} corpus reaches up to 69.16\%, which is the best performance on that corpus. However, using the single method of the Jamo subword unit resulted in a relatively low performance of 95.19\% on the \textbf{WF} corpus due to the lack of pre-trained embeddings. By using the Jamo subword unit together with other subword units such as morphemes or BPE units, the system was able to achieve excellent performance on the \textbf{WF} corpus alike.

Additionally, $\textbf{IEE}_\textbf{C}[mbj]\textbf{+ALL}$ achieves the best performance of 97.47\% on the \textbf{WF} corpus by combining the morpheme embedding vectors with BPE embedding vectors. The result suggests that one can expect additional system performance improvement by further integrating different types of pre-trained subword embeddings. The evaluation result also shows that the Jamo subword unit is more effective compared to the character subword unit in terms of the \textbf{KM} corpus performance. Considering that Korean users are more likely to type in \textit{Jamo}s than characters, this result is quite understandable.

\subsection{Error Analysis}
Several examples are observed to figure out why the proposed Eojeol-based approach works better than the existing morpheme-based approaches. Those examples are presented in table \ref{tab:comp_ex}; important clues for sentence classification are marked in bold.

\begin{table}
\centering
\begin{tabular}{|l|l|l|} \hline
\multicolumn{3}{|l|}{\textbf{Case 1.} Is the \textbf{light turned on} now?} \\ \hline
& \textbf{S} & 지금 \textbf{불}링 \textbf{켜}졌어? \\ \cline{2-3}
\textbf{T}& \textbf{M} &  지금 불링 \textbf{켜다}(turned on)/지다/었/어 \\ \cline{2-3}
& \textbf{B} & \_지금 \_\textbf{불}/링 \_\textbf{켜}(turned on)/졌/어 \\ \hline

& \textbf{S} &  지금 \textbf{불}이 \textbf{켜}졌어? \\ \cline{2-3}
\textbf{C}& \textbf{M} & 지금 \textbf{불}/이 \textbf{켜다}/지다/었/어 \\ \cline{2-3}
& \textbf{B} & \_지금 \_불이 \_\textbf{켜}/졌/어 \\ \hline

\multicolumn{3}{|l|}{\textbf{Case 2.} \textbf{Lottery number}.} \\ \hline
& \textbf{S} & \textbf{로또버ㄴ호}\\ \cline{2-3}
\textbf{T}&  \textbf{M} &로또버/ㄴ/호\\  \cline{2-3}
& \textbf{J} & ㄹㅗㄸㅗㅂㅓㄴㅎㅗ\\ \hline
& \textbf{S} & \textbf{로또번호} \\ \cline{2-3}
\textbf{C}&  \textbf{M} &\textbf{로또(lottery)}/\textbf{번호(number)} \\  \cline{2-3}
& \textbf{J} & ㄹㅗㄸㅗㅂㅓㄴㅎㅗ \\ \hline
\end{tabular}
\caption{Examples for which the Eojeol-based approach works better than the morpheme-based approaches. In the first column, \textbf{T} means the sentence with typo, and \textbf{C} means the corrected sentences; in the second column, \textbf{S}, \textbf{M}, \textbf{B} and \textbf{J} means sentence, morpheme subunits, BPE subunits and Jamo subunits, respectively. }
 
\label{tab:comp_ex}
\end{table}

For \textbf{Case 1 (typo\textbf{T})}, the vital clue ``\textbf{불}(light)" is not extracted as morphemes due to spelling error. However, the clue is successfully recovered in BPE subword units. For the \textbf{Case  2}, the typo can be handled by considering the Jamo subword units. The Jamo subword unit lists for correct and wrong sentences are the same, while their morpheme and BPE subword unit lists are different. As observed in the examples, the system can get a clue from other types of subword units by integrating multiple different types of subword unit embeddings when one subword unit type fails to recover the vital clue.

The proposed algorithm still has its weakness in redundantly spaced sentences, which is exemplified by the sentence ``ㅈ ㅗ명 꺼줘". It is the misspelling of ``조명 꺼줘(Turn off the light)". In the example sentence, the vital clue ``조명(light)" is separated into two different Eojeols ``ㅈ" and ``ㅗ명" due to the misplaced space. Since they are separated into two Eojeols, the proposed Eojeol-based algorithm fails to recover the critical clue ``조명" and to get the correct classification result.

\section{Conclusion}
\label{sec:con}

In this paper, a novel approach of Integrated Eojeol Embedding is proposed to handle the Korean erroneous sentence classification tasks. Two noise insertion methods are additionally proposed to overcome the weakness of the Eojeol-embedding-based approaches and to add noises into training data automatically. The proposed system is evaluated against the intent classification corpus for Korean chatbot. The evaluation result shows over 18\%p improvement on the erroneous sentence classification task and 0.5\%p improvement on the grammatically correct sentence classification task, compared to the baseline system.  

Although the proposed algorithm is tested only against the Korean chatbot intent classification task, it can be applied to other types of sentence classification, such as sentiment analysis or comment categorization. Also, the application of the proposed algorithm need not be restricted to the Korean text. For example, it can be applied to English text to integrate the English GloVe embedding vectors and BPE unit embedding vectors.

Our next work will be to investigate the performance of the proposed algorithm further and expand the algorithm to cover other languages.

\bibliography{ie}

\begin{thebibliography}{17}
\expandafter\ifx\csname natexlab\endcsname\relax\def\natexlab#1{#1}\fi

\bibitem[{Ba et~al.(2016)Ba, Kiros, and Hinton}]{Ba16}
Jimmy~Lei Ba, Jamie~Ryan Kiros, and Geoffrey~E. Hinton. 2016.
\newblock Layer normalization.
\newblock \emph{Computing Research Repository}, arXiv:1607.06450.

\bibitem[{Bowman et~al.(2015)Bowman, Angeli, Potts, and Manning}]{Bowman15}
Samuel~R. Bowman, Gabor Angeli, Christopher Potts, and Christopher~D. Manning.
  2015.
\newblock A large annotated corpus for learning natural language inference.
\newblock In \emph{Proceedings of the 2015 Conference on Empirical Methods in
  Natural Language Processing (EMNLP)}.

\bibitem[{Choi et~al.(2018)Choi, Park, Lim, Baek, Lee, Shin, Kim, and
  Shin}]{Choi18}
DongHyun Choi, IlNam Park, Jae-Soo Lim, SeulYe Baek, MiOk Lee, Myeongcheol
  Shin, EungGyun Kim, and Dong~Ryeol Shin. 2018.
\newblock Sentence classification for korean dialog engine.
\newblock In \emph{Proceedings of HCLT}, pages 210--214.

\bibitem[{Chollet(2017)}]{Chollet17}
Francois Chollet. 2017.
\newblock Xception: Deep learning with depthwise separable convolutions.
\newblock In \emph{IEEE Conference on Computer Vision and Pattern
  Recognition(CVPR)}, pages 1800--1807. IEEE.

\bibitem[{Devlin et~al.(2019)Devlin, Chang, Lee, and Toutanova}]{bert}
Jacob Devlin, Ming-Wei Chang, Kenton Lee, and Kristina Toutanova. 2019.
\newblock Bert: Pre-training of deep bidirectional transformers for language
  understanding.
\newblock In \emph{Proceedings of the 2019 Conference of the North American
  Chapter of the Association for Computational Linguistics: Human Language
  Technologies}, volume~1, pages 4171--4186.

\bibitem[{Heinzerling and Strube(2018)}]{heinzerling2018bpemb}
Benjamin Heinzerling and Michael Strube. 2018.
\newblock {BPEmb: Tokenization-free Pre-trained Subword Embeddings in 275
  Languages}.
\newblock In \emph{Proceedings of the Eleventh International Conference on
  Language Resources and Evaluation (LREC 2018)}, pages 2990--2993.

\bibitem[{Im and Cho(2017)}]{Im17}
Jinbae Im and Sungzoon Cho. 2017.
\newblock Distance-based self-attention network for natural language inference.
\newblock \emph{Computing Research Repository}, arXiv:1712.02047.

\bibitem[{Kim(2014)}]{Kim14}
Yoon Kim. 2014.
\newblock Convolutional neural networks for sentence classification.
\newblock In \emph{Proceedings of the 2014 Conference on Empirical Methods in
  Natural Language Processing}, pages 1746--1751.

\bibitem[{Kingma and Ba(2015)}]{Kingma14}
Diederik~P. Kingma and Jimmy Ba. 2015.
\newblock Adam: A method for stochastic optimization.
\newblock In \emph{International Conference on Learning Representations},
  volume~5.

\bibitem[{Oh et~al.(2017)Oh, Lee, Choi, and Hur}]{Hur17}
Kyo-Joong Oh, Dongkeon Lee, Ho-Jin Choi, and Jeong Hur. 2017.
\newblock A model of korean sentence similarity measurement using sense-based
  morpheme embedding and rnn sentence encoding.
\newblock In \emph{2017 IEEE International Conference on Big Data and Smart
  Computing}, pages 430--433.

\bibitem[{Park et~al.(2018{\natexlab{a}})Park, Choi, Shin, and Kim}]{Park18_2}
IlNam Park, DongHyun Choi, MyeongCheol Shin, and EungGyun Kim.
  2018{\natexlab{a}}.
\newblock Korean sentence classification system using glove and maximum entropy
  model.
\newblock In \emph{Proceedings of HCLT}, pages 522--526.

\bibitem[{Park et~al.(2018{\natexlab{b}})Park, Kim, and Kang}]{Park18}
Keunyoung Park, Kyungduk Kim, and Inho Kang. 2018{\natexlab{b}}.
\newblock Jam-packing korean sentence classification method robust for spacing
  errors.
\newblock In \emph{Proceedings of HCLT}, pages 600--604.

\bibitem[{Pennington et~al.(2014)Pennington, Socher, and
  Manning}]{pennington14}
Jeffrey Pennington, Richard Socher, and Christopher~D. Manning. 2014.
\newblock \href {http://www.aclweb.org/anthology/D14-1162} {Glove: Global
  vectors for word representation}.
\newblock In \emph{Empirical Methods in Natural Language Processing (EMNLP)},
  pages 1532--1543.

\bibitem[{Srivastava et~al.(2014)Srivastava, Hinton, Krizhevsky, Sutskever, and
  Salakhutdinov}]{Srivastava14}
Nitish Srivastava, Geoffrey Hinton, Alex Krizhevsky, Ilya Sutskever, and Ruslan
  Salakhutdinov. 2014.
\newblock Dropout: a simple way to prevent neural networks from overfitting.
\newblock \emph{The Journal of Machine Learning Research}, 15(1):1929--1958.

\bibitem[{Yin and Sch{\"u}tze(2015)}]{Yin16}
Wenpeng Yin and Hinrich Sch{\"u}tze. 2015.
\newblock Multichannel variable-size convolution for sentence classification.
\newblock In \emph{Proceedings of the Nineteenth Conference on Computational
  Natural Language Learning}, pages 204--214.

\bibitem[{Yoon et~al.(2018)Yoon, Lee, and Lee}]{Yoon18}
Deunsol Yoon, Dongbok Lee, and Sangkeun Lee. 2018.
\newblock Dynamic self-attention : Computing attention over words dynamically
  for sentence embedding.
\newblock \emph{Computing Research Repository}, arXiv:1808.073837.

\bibitem[{Zhou et~al.(2015)Zhou, Sun, Liu, and Lau}]{Zhou15}
Chunting Zhou, Chonglin Sun, Zhiyuan Liu, and Francis~C.M. Lau. 2015.
\newblock A c-lstm neural network for text classification.
\newblock \emph{Computing Research Repository}, arXiv:1511.08630.

\end{thebibliography}
\bibliographystyle{acl_natbib}

\end{document}